\documentclass{bmvc2k}

\title{Multi-Source domain adaptation via supervised contrastive learning and confident consistency regularization}

\addauthor{Marin Scalbert}{marin.scalbert@centralesupelec.fr}{1, 2}
\addauthor{Maria Vakalopoulou}{maria.vakalopoulou@centralesupelec.fr}{1}
\addauthor{Florent Couzini\'e-Devy}{f.couzinie-devy@vitadx.com}{2}

\addinstitution{
 MICS\\
 CentraleSup\'elec\\
 Gif-sur-Yvette, France
}
\addinstitution{
 VitaDX International\\
 Paris, France
}

\runninghead{Scalbert, Vakalopoulou, Couzini\'e-Devy}{CMSDA}

\usepackage{amssymb}
\usepackage{svg}
\usepackage{amsbsy, bm}
\usepackage{bbm}
\usepackage{float}
\usepackage{pifont}
\usepackage{tikz}
\usepackage{caption}
\usepackage{float}
\usepackage{hyperref}
\usepackage{adjustbox}
\newcommand{\card}[1]{\vert #1 \vert}

\begin{document}

\setlength\abovecaptionskip{-0.7\baselineskip}
\maketitle

\begin{abstract}
Multi-Source Unsupervised Domain Adaptation (multi-source UDA) aims to learn a model from several labeled source domains while performing well on a different target domain where only unlabeled data are available at training time.
To align source and target features distributions, several recent works use source and target explicit statistics matching such as features moments or class centroids. Yet, these approaches do not guarantee class conditional distributions alignment across domains. In this work, we propose a new framework called Contrastive Multi-Source Domain Adaptation (CMSDA) for multi-source UDA that addresses this limitation.
Discriminative features are learned from interpolated source examples via cross entropy minimization and from target examples via consistency regularization and hard pseudo-labeling. Simultaneously, interpolated source examples are leveraged to align source class conditional distributions through an interpolated version of the supervised contrastive loss. This alignment leads to more general and transferable features which further improves the generalization on the target domain. Extensive experiments have been carried out on three standard multi-source UDA datasets where our method reports state-of-the-art results.
\end{abstract}


\section{Introduction}
The performances of deep learning models are known to drop when  training and testing data have different distributions. This phenomenon, known as \textit{Domain Shift}, has led to the emergence of the Unsupervised Domain Adaptation (UDA) problem that has been intensively studied over the recent years~\cite{RTN, JAN, DAN, deep_domain_confuzion, deep_hashing_network, DTN, CORAL, xie_learning_nodate, CAN, cluster_alignement_da}.
Single-source UDA aims to leverage labeled examples from a source domain and unlabeled examples from a target domain to learn a model that performs well on unseen target examples.
In the most practical scenario, to collect as much labeled data as possible, several source domains are considered rather than a single one. In such case, the setting is referred to as multi-source UDA.

To solve the UDA problem, most of the methods learn discriminative features from labeled source data and exploit unlabeled target data to align source and target distributions. Source and target distributions alignment is performed so as to maintain the discriminative power of the model on the target domain. Plethora of UDA methods, in the single-source~\cite{RTN, JAN, DAN, deep_domain_confuzion, deep_hashing_network, DTN} or multi-source settings~\cite{DCTN, MDAN, M3SDA} have tried to align source and target marginal distributions. However, these methods are susceptible to fail if source and target class conditional distributions are not aligned~\cite{uda_conditional_alignment}. Alignment of source and target class conditional distributions can be achieved through adversarial based methods such as~\cite{uda_conditional_alignment} but they tend to be cumbersome to train while the alignment of the domains can fail if pseudo labels on target examples are noisy.
To align source and target class conditional distributions,~\cite{xie_learning_nodate} has proposed to match source and target class centroids. Nevertheless, in order to estimate accurately the true class centroids, batches should be carefully designed to contain enough examples per class while maintaining a well-tuned moving average of centroids. This method assumes also that a single centroid can represent the whole distribution in a class which is a wrong assumption in case of multimodal class distributions. 

In this work, to address the problem of multi-source UDA and align efficiently source class conditional distributions, we introduce a new framework named Contrastive Multi-Source Domain Adaptation (CMSDA). CMSDA learns discriminative features on source examples via cross entropy minimization and align class conditional distributions of all source domains through supervised contrastive loss. Source class conditional distributions alignment leads to more general and transferable features for the target domain. In the same time, the model adjusts to the target domain via hard pseudo labeling and consistency regularization. To further enhance the robustness, calibration of our model and enable deeper exploration of each source domain input space, MixUp~\cite{mixup_training} is leveraged on source domains. However, MixUp could be replaced by any other mixing methods such as CutMix \cite{cutmix}. Interpolating source examples from different source domains can even be seen as way to mix source domains styles. 
Since MixUp is performed on source domains, interpolated versions of the cross entropy and supervised contrastive losses are used in the final objective rather than the standard versions.
To sum up, our contributions are the following: \textbf{(1)} we design a novel tailored end-to-end architecture that maps the different domains to a common latent space, and efficiently transfers knowledge learned on source domains to the target domain using 
recent advances of supervised contrastive learning, semi-supervised learning and mixup training; 
\textbf{(2)} we show for the first time that supervised contrastive learning and its interpolated extension can be used in the context of domain adaptation for source class conditional distributions alignment leading to higher accuracy for target domains with large domain shift, \textbf{(3)} we report state of the art results on three standard multi-source UDA datasets.


\section{Related works}
\label{sec:related_works}
\textbf{Multi-source UDA. }
In the multi-source UDA setting, former methods such as MDAN~\cite{MDAN} have tried to extend a single-source UDA method to the multi-source setting. In MDAN, each source and target marginal distributions are aligned via an adversarial based method ~\cite{gradient_reversal_layer}.
In DCTN~\cite{DCTN}, an adversarial based method is proposed to align marginally each source with the target and perplexity scores, measuring the possibilities that a target sample belongs to the different source domains, are used to weight predictions of different source classifiers. M$^{3}$SDA-$\beta$~\cite{M3SDA} uses a two-steps approach combining an ensemble of source classifiers. In the first step, the method aligns marginally each source with the target domain and also each pair of source domains by matching first order moments of features maps channels. In the second step, to enhance distributions alignment, the different source classifiers are trained following an adversarial method~\cite{MCD}. CMSS~\cite{CMSS} exploits an original adversarial approach that selects dynamically the source domains and examples that are the most suitable for aligning source and target distributions. DAEL~\cite{DAEL} combines a collaborative training of an ensemble of source expert classifiers with hard pseudo labeling and consistency regularization on the target domain. For source examples, robust features are learned by ensuring consistency between the expert source classifier and the ensemble of non-expert source classifiers. For unlabeled target examples, since no expert target classifier is available, consistency is ensured between the most confident source expert classifier and the ensemble of other source expert classifiers. Our method shares some similarities with DAEL as it exploits similar semi-supervised learning techniques to learn on unlabeled target data however, ours adds an additional constraint to align source class conditional distributions. 
Recent works are also focusing on improving UDA methods by using specific data augmentation. For instance, MixUp has been explored for single source UDA methods ~\cite{dual_mixup, adversarial_da_mixup} but the literature on methods exploiting MixUp for problems such as multi-source UDA and multi-source Domain Generalization is still very sparse \cite{dg_mixup, cumix}. Our method exploits MixUp but also an interpolated version of the supervised contrastive loss to work with the soft labels produced by this data augmentation.
Finally, multi-source UDA methods have also explored other types of approaches \cite{MFSAN, MDMN, MDDA, LtC-MSDA}, different setting \cite{moe} or other data modalities \cite{mosdanet}.  

\textbf{Contrastive learning.} Recently, representation learning has known major breakthroughs due to the advance in the field of contrastive learning~\cite{CPC, SimCLR, MOCO, BYOL, SWaV, SimSiam}. The main idea behind most of contrastive learning methods is that similar examples should share the same representation. 
For example, in SimCLR~\cite{SimCLR} the loss enforces pairs of augmented versions of the same image (positives) to have the same representation while having dissimilar representations from all other examples in the batch (negatives).
Most recently, the loss used in~\cite{SimCLR} has been extended for the supervised setting~\cite{supervised_contrastive_learning}. With the supervised contrastive loss, examples belonging to the same class are pushed closer while examples from different classes are pushed apart.
There have been some attempts to adapt contrastive learning in the case of single-source UDA~\cite{CAN}. However, supervised contrastive learning has yet to be explored on the multi-source UDA problem and seems a natural and an appropriate approach to align source class conditional distributions.

\textbf{Semi-supervised learning.} To exploit unlabeled examples, common semi-supervised approaches are based either on consistency regularization~\cite{mean_teachers, MixMatch, UDA, transformation_consistency_regularization} or pseudo labeling~\cite{pseudo_labeling}.
Consistency regularization learns on unlabeled data by relying on the assumption that the model should output similar predictions when perturbed versions of the same image are presented. Hard pseudo-labeling consists of using hard predictions on unlabeled examples as ground truth labels for these examples.
Some semi-supervised methods such as FixMatch~\cite{FixMatch} uses both approaches. In FixMatch, strongly and weakly augmented images are generated from the same unlabeled image. The network is then trained to ensure consistency between the prediction on the strongly augmented image and the hard pseudo label obtained on  the weakly augmented image. To handle potential false pseudo labels, only confident pseudo-labeled examples contributes to the loss. In our method, these two different kind of approaches are leveraged to learn on the unlabeled target data. 

\section{CMSDA Framework}
\label{sec:method}
{
    \begin{figure}[htbp]
    	\begin{center}
    	\includegraphics[width=1\linewidth]{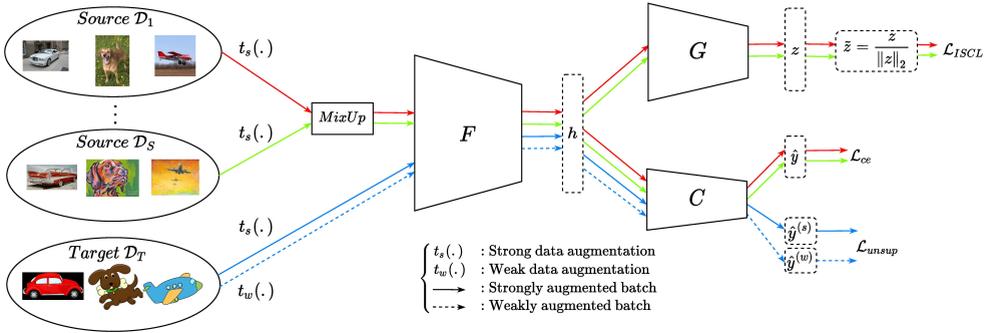}
    	\end{center}
     	\caption{\textbf{CMSDA framework.} CMSDA contains 3 shared components: a features extractor $F$, a projection head $G$ and a classification head $C$. Source examples are strongly augmented via $t_s(.)$ and interpolated using MixUp. Target examples are weakly and strongly augmented via $t_w(.)$ and  $t_s(.)$. All examples are fed to $F$ to produce the embeddings $\bm{h}$. In the top branch, source embeddings $\bm{h}$ are given to $G$ to produce the representations $\bm{\tilde{z}}$ which are used to align source class conditional distributions via minimization of $\mathcal{L}_{ISCL}$. In the bottom branch, all embeddings $\bm{h}$ are fed to $C$ to produce probability vectors $\bm{\hat{y}}$, $\bm{\hat{y}^{(s)}}$ and $\bm{\hat{y}^{(w)}}$. These probability vectors are used to learn discriminative features via minimization of $\mathcal{L}_{ce}$ and $\mathcal{L}_{unsup}$.}
    \label{fig:model}
    \end{figure}
}
In the setting of multi-source UDA, we are given $S$ different source domains $\{\mathcal{D}_1, \hdots, \mathcal{D}_S\}$ and one target domain $\mathcal{D}_T$. Each of the source domain $\mathcal{D}_i$ contains $n_{\mathcal{D}_i}$ labeled examples $\{(\bm{x_j}, \bm{y_j}) \mid 1 \leq j \leq n_{\mathcal{D}_i}\}$ while target domain contains $n_{\mathcal{D}_T}$ unlabeled examples $\{\bm{x_j} \mid 1 \leq j \leq n_{\mathcal{D}_T}\}$. The goal of multi-source UDA is to learn a robust model from the $S$ labeled source domains $\mathcal{D}_1, \ldots, \mathcal{D}_S$ and the target domain $\mathcal{D}_T$ so that it generalizes well on unseen target domain examples.

\subsection{Model architecture}
Our model architecture (\autoref{fig:model}) is shared for source and target domains. It is composed of three different parts:

\textbf{Features extractor.} The features extractor $F$ is a convolutional neural network. It takes an input image $\bm{x} \in \mathbb{R}^{H \times W \times 3}$ and returns a features vector $\bm{h} \in \mathbb{R}^{d_1}$.

\textbf{Projection head.} The projection head $G$ takes as input the representation $\bm{h}$ and outputs a lower dimensional representation $\bm{z} \in \mathbb{R}^{d_2}$ with $d_1 > d_2$. Similarly to~\cite{supervised_contrastive_learning}, $G$ is a multi-layer perceptron consisting in two fully connected layers. The first layer preserves the dimension while the second reduces the dimensions from $d_1$ to $d_2$. Previous self-supervised contrastive learning methods~\cite{MOCO, BYOL, SWaV, SimSiam} indicate that the use of a batch normalization layer after the first fully connected layer has shown to generate more powerful representations. Following these findings, we include a batch normalization after the first fully connected layer of $G$.

\textbf{Classification head.} The classification head $C$ is responsible for the final predictions. In previous self-supervised~\cite{SimCLR} or supervised~\cite{supervised_contrastive_learning} contrastive learning methods, the projection head is usually removed after the training and a linear classifier is fine-tuned on top of the frozen representation $\bm{h}$. Indeed, in practice, $\bm{h}$ provides better representations than $\bm{z}$ for the final classification task. In this study, we follow the same strategy but still investigate this design choice in the experiments. Therefore, $C$ is a single fully connected layer that takes as input the features extractor representation $\bm{h}$ and outputs a probability vector $\bm{\hat{y}} \in [0,1] ^ {K}$ where $K$ is the number of classes.
\subsection{Optimization Strategy}
\textbf{Source domains interpolation with MixUp. }
MixUp ~\cite{mixup_training} performs data augmentation by creating new examples $(\bm{\tilde{x}}, \bm{\tilde{y}})$ as convex combinations of random pairs of examples and their corresponding labels $(\bm{x_a}, \bm{y_a})$ and $(\bm{x_b} , \bm{y_b})$:
\begin{equation}
    \begin{cases}
        \bm{\tilde{x}} = \lambda \bm{x_a} + (1-\lambda)\bm{x_b}\\
        \bm{\tilde{y}} = \lambda \bm{y_a} + (1-\lambda)\bm{y_b}
    \end{cases}
\end{equation}
where $\lambda \sim Beta(\alpha, \alpha)$. In CMSDA, MixUp is applied on source examples right after strong data augmentation $t_s(.)$.

\textbf{Overall objective.} The overall objective includes three different losses: the interpolated cross entropy loss $\mathcal{L}_{ce}$, the interpolated supervised contrastive loss $\mathcal{L}_{ISCL}$ and the unsupervised FixMatch loss $\mathcal{L}_{unsup}$. The final objective minimized by the model can be written:
\begin{equation}
	\mathcal{L} = \mathcal{L}_{ce} + \lambda_{1}\mathcal{L}_{ISCL} + \lambda_{2}\mathcal{L}_{unsup}
	\label{eq:overall_objective}
\end{equation}
$\lambda_{1}$ and $\lambda_{2}$ are hyperparameters weighting the contributions of the losses $\mathcal{L}_{ISCL}$ and $\mathcal{L}_{unsup}$.

\textbf{Interpolated cross-entropy.} In order to leverage the interpolated labeled source examples obtained after MixUp, our framework minimizes an interpolated cross entropy denoted $\mathcal{L}_{ce}$. For a single interpolated source example $(\bm{\tilde{x_i}}, \bm{\tilde{y_i}}) = (\lambda \bm{x_a} + (1-\lambda) \bm{x_b}, \lambda \bm{y_a} + (1-\lambda) \bm{y_b})$ with $\bm{\hat{y}}$ the prediction on the example $\bm{\tilde{x_i}}$, the interpolated cross entropy per sample denoted $\mathcal{L}_{i}^{ce}$ can be written:
\begin{equation}
    \mathcal{L}_{i}^{ce} = H(\lambda \bm{y_a} + (1-\lambda) \bm{y_b}, \bm{\hat{y}})
\end{equation}
$H(\bm{p}, \bm{q})$ denotes the cross entropy between a reference distribution $\bm{p}$ and an approximated distribution $\bm{q}$. 
$\mathcal{L}_{ce}$ is then computed by averaging the loss per sample on $N_S$ interpolated source examples.

\textbf{Interpolated supervised contrastive loss.} To align source class conditional distributions, supervised contrastive loss (SCL) seems a simple and straightforward option. When minimizing SCL, representations of examples belonging to the same class are pulled together while representations of examples belonging to different classes are pushed away. In the case of multiple source domains, SCL would force learned features to be domain invariant.
Given an example $(\bm{x_i}, \bm{y_i})$ with normalized projection head representation $\bm{\tilde{z_i}}$, the per sample SCL is defined by:
\begin{equation}
    \resizebox{0.90\hsize}{!}{%
	 $\mathcal{L}^{SCL}(\bm{\tilde{z_i}}, \bm{y_i}) = \displaystyle - \dfrac{1}{\card{P(i, \bm{y_i})}} \sum_{p \in P(i, \bm{y_i})}^{} \log{\left (\dfrac{e^{\dfrac{\bm{\tilde{z}}_i.\bm{\tilde{z}}_p}{T}}}{\displaystyle \sum_{j \in A(i)}e^{\dfrac{\bm{\tilde{z}}_i.\bm{\tilde{z}}_j}{T}}} \right)} \
	\text{where: }
	\begin{cases}
		A(i) = \{1, \ldots, N_S\} \setminus \{i\}\\
		P(i, \bm{y_i}) = \{j \in A(i) \mid  \bm{y_j} = \bm{y_i}\}
	\end{cases}$
	}
	\label{eq:supervised_contrastive_loss}
\end{equation}
Here, $T$ corresponds to a temperature hyperparameter, $A(i)$ stands for the anchors set (indexes of examples different than $i$) and $P(i, \bm{y_i})$ the positives set (indexes of other examples whose label is equal to $\bm{y_i}$).

However, by using MixUp on source examples, we end up with examples with soft labels whereas SCL requires hard labels. Indeed, SCL needs hard labels so that examples with same labels can be identified and be pulled together. To circumvent the soft labels problem raised by MixUp, we instead use an interpolated version of SCL (ISCL) introduced in~\cite{interpolated_supervised_contrastive_loss} and minimize it on $N_S$ interpolated source examples.

Given an interpolated source example $(\bm{\tilde{x_i}}, \bm{\tilde{y_i}}) = (\lambda \bm{x_a} + (1-\lambda) \bm{x_a}, \lambda \bm{y_a} + (1-\lambda) \bm{y_b})$ with $\bm{\tilde{z_i}}$ the normalized projection head representation of the example $\bm{\tilde{x_i}}$, the per sample ISCL denoted $\mathcal{L}_{i}^{ISCL}$ is defined by:
\begin{equation}
    \mathcal{L}_{i}^{ISCL} = \lambda \mathcal{L}^{SCL}(\bm{\tilde{z_i}}, \bm{y_a}) + (1-\lambda) \mathcal{L}^{SCL}(\bm{\tilde{z_i}}, \bm{y_b})
\end{equation}
$\mathcal{L}^{SCL}(\bm{\tilde{z_i}}, \bm{y_a})$ ($\mathcal{L}^{SCL}(\bm{\tilde{z_i}}, \bm{y_b})$) stands for the per sample SCL for the example $\bm{x_i}$ with label $\bm{y_a}$ ($\bm{y_b}$). The definition of $P(i, \bm{y_i})$ in Equation \ref{eq:supervised_contrastive_loss} implies that the examples in $A(i)$ have hard labels. Therefore, as in~\cite{interpolated_supervised_contrastive_loss}, for each interpolated example in $A(i)$, we consider as hard label the dominant label which is the one associated to the highest mixing coefficients $(\lambda, 1-\lambda)$. Finally, $\mathcal{L}_{ISCL}$ is computed by averaging the loss per sample on $N_S$ interpolated source examples.

\textbf{Consistency regularization and hard-pseudo labeling.}
In our framework, given $N_T$ unlabeled examples drawn from the target domain, we apply a weak augmentation $t_w(.)$ and a strong augmentation $t_s(.)$ on each example to obtain a weakly augmented example $\bm{x^{(w)}}$ and strongly augmented example $\bm{x^{(s)}}$. $\bm{x^{(w)}}$ and $\bm{x^{(s)}}$  are fed to $F$ and $C$ to obtain respectively the predictions $\bm{\hat{y}^{(w)}}$ and $\bm{\hat{y}^{(s)}}$. The hard prediction of the weakly augmented example denoted $\arg\max \bm{\hat{y}^{(w)}}$ \footnote{Similar to~\cite{FixMatch}, for simplicity, we assume that $\arg\max$ applied on a $K$ dimensional probability vector gives a valid $K$ dimensional one-hot vector.} is used as a hard pseudo label while we ensure consistency between the predictions on the strongly example $\bm{\hat{y}^{(s)}}$ and the hard pseudo label of the weakly augmented example $\arg\max \bm{\hat{y}^{(w)}}$. As described in Section \ref{sec:related_works}, to discard potential noisy pseudo labels, only pseudo-labeled weakly augmented examples with a maximum predicted probability above some fixed probability threshold $\tau$ contributes to the loss. This corresponds to minimizing the unsupervised loss term of ~\cite{FixMatch} defined by:
\begin{equation}
	\mathcal{L}_{unsup} = \displaystyle \dfrac{1}{N_T} \sum_{i=1}^{N_T} \mathbbm{1}_{\{\max \bm{\hat{y}^{(w)}} > \tau\}}H\left (\arg\max \bm{\hat{y}^{(w)}} , \bm{\hat{y}^{(s)}} \right )
\end{equation}
\section{Experiments}
\subsection{Evaluation}
We evaluate and compare our method on three 
standard multi-source UDA datasets: DomainNet, MiniDomainNet and Office-Home. For each dataset and target domain, two standard baselines commonly used in the context of multi-source UDA~\cite{DCTN, M3SDA, CMSS, DAEL} have been added: \textit{Source-only} and \textit{Oracle}. \textit{Source-only} represents a model trained only on source examples with standard cross-entropy whereas \textit{Oracle} represents a model trained with labeled target examples. Performances on MiniDomainNet and DomainNet are averaged over three runs with different random seeds. The performances of our method along with the compared multi-source UDA methods for the datasets DomainNet, MiniDomainNet and Office-Home are respectively reported on \autoref{tab:perfs_domain_net},  \autoref{tab:perfs_mini_domain_net} and \autoref{tab:perfs_office_home}. For easier interpretation, first and second best methods are respectively highlighted in \textcolor{red}{\textbf{bold red}} and \textit{\textcolor{blue}{italic blue}}. Datasets information and implementation details can be found in the supplementary material.

\textbf{DomainNet. }
Our method achieves the best performance with $50.42\%$ average accuracy which is $+1.72\%$ gain over previous state of the art. Overall, our method reports the best performances on 4 out of 6 target domains and the second best performance on one of the two other target domains. 
On the \textit{Quickdraw} domain, our framework achieves the best performance by a large margin $(+2.40\%)$ compared to the second best method (SImpAl). On this challenging target domain, SImpaL, CMSS, LtC-MSDA and our method are the only ones that are not subject to negative transfer~\cite{negative_transfer} (lower performances than the baseline \textit{Source-only}). This indicates that our method is able to work even for complex target domains.

\textbf{MiniDomainNet. }
Our method achieves the best overall accuracy with $61.90\%$, the best accuracy on 3 out of 4 target domains and the second best on the last domain.

\textbf{Office-Home. }
Our method achieves the best overall accuracy with $76.60\%$. More specifically, CMSDA obtains the best/second best accuracies on 2 out of 4 target domains and competitive results on the two other target domains.

\begin{center}
    \captionsetup{type=table}
    \begin{tabular}{c}
        \begin{minipage}{0.8\textwidth}
            \begin{center}
                \begin{adjustbox}{width=1\textwidth,center}
                    \begin{tabular}{|l|cccccc|c|}
                        \hline
                        & \multicolumn{6}{|c|}{Target domain} &\\
                        \hline
                        Methods & \textit{Clp} & \textit{Inf} & \textit{Pnt} & \textit{Qdr} & \textit{Rel} & \textit{Skt} & \textit{Avg} \\
                        \hline
                        \textit{Source-only}~\cite{DAEL} & $47.60\pm0.52$ & $13.00\pm0.41$ & $38.10\pm0.45$ & $13.30\pm0.39$ & $51.90\pm0.85$ & $33.70\pm0.54$ & $32.90$\\
                        \textit{Oracle}~\cite{DAEL} & $69.30\pm 0.37$ & $34.50\pm0.42$ & $66.30\pm0.67$ & $66.80\pm0.51$ & $80.10\pm0.59$ & $60.70\pm0.48$ & $63.00$ \\
                        \hline
                        DANN~\cite{DANN} & $45.50\pm 0.59$ & $13.10 \pm 0.41$ & $37.00 \pm 0.69$ & $13.20 \pm 0.77$ & $48.90\pm 0.65$ & $31.80\pm 0.62$ & $32.60$ \\
                        DCTN~\cite{DCTN} & $48.60\pm 0.73$ & $23.50 \pm 0.59$ & $48.80 \pm 0.63$ & $7.20 \pm 0.46$ & $53.50 \pm 0.56$ & $47.30 \pm 0.47$ & $38.20$ \\
                        MCD~\cite{MCD} & $54.30 \pm 0.64$ & $22.10 \pm 0.70$ & $45.70 \pm 0.63$ & $7.60 \pm 0.49$ & $58.40 \pm 0.65$ & $43.50 \pm 0.57$ & $38.50$ \\
                        M$^3$SDA-$\beta$~\cite{M3SDA} & $58.60 \pm 0.53$ & $26.00 \pm 0.89$ & $52.30 \pm 0.55$ & $6.30 \pm 0.58$ & $62.27 \pm 0.51$ & $49.50 \pm 0.76$ & $42.60$ \\
                        CMSS~\cite{CMSS} & $64.20 \pm 0.18$ & \textcolor{blue}{$\mathit{28.00 \pm 0.20}$} & $53.60 \pm 0.39$ & $16.00 \pm 0.12$ & $63.40 \pm 0.21$ & $53.80 \pm 0.35$ & $46.50$ \\
                        LtC$-$MSDA~\cite{LtC-MSDA} & $63.10 \pm 0.50$ & \textcolor{red}{$\bm{28.70 \pm 0.70}$} & $56.10 \pm 0.50$ & $16.30 \pm 0.50$ & $66.10 \pm 0.60$ & $53.80 \pm 0.60$ & $47.40$ \\
                        SImpAl$_{101}$~\cite{SImpAl} & $66.40 \pm 0.80$ & $26.50 \pm 0.50$ & $56.60 \pm 0.70$ & \textcolor{blue}{$\mathit{18.90 \pm 0.80}$} & \textcolor{blue}{$\mathit{68.00 \pm 0.50}$} & $55.50 \pm  0.30$ & $48.60$ \\
                        DAEL~\cite{DAEL} & \textcolor{blue}{$\mathit{70.80 \pm 0.14}$} & $26.50 \pm 0.13$ & \textcolor{blue}{$\mathit{57.40 \pm 0.28}$} & $12.20 \pm0.70$ & $65.00 \pm 0.23$ & \textcolor{red}{$\bm{60.60 \pm  0.25}$} & \textcolor{blue}{$\mathit{48.70}$} \\
                        Ours & \textcolor{red}{$\bm{70.95\pm0.23}$} & $26.58\pm0.34$ & \textcolor{red}{$\bm{57.56\pm0.08}$} & \textcolor{red}{$\bm{21.30\pm0.11}$} & \textcolor{red}{$\bm{68.12 \pm 0.22}$} & \textcolor{blue}{$\mathit{59.48\pm0.07}$} & \textcolor{red}{$\bm{50.42}$} \\
                        \hline
                   \end{tabular}
                \end{adjustbox}
            \end{center}
        \captionof{table}{Accuracy (\%) on DomainNet (\textit{Clp}: \textit{Clipart}, \textit{Inf}: \textit{Infograph}, \textit{Pnt}: \textit{Painting}, \textit{Qdr}: \textit{Quickdraw}, \textit{Rel}: \textit{Real}, \textit{Skt}: \textit{Sketch}, \textit{Avg}: \textit{Average}).}
        \label{tab:perfs_domain_net}
        \end{minipage}
        \\
        \begin{tabular}{cc}
            \begin{minipage}{0.56\textwidth}
                \begin{center}
                \begin{adjustbox}{width=1\textwidth,center}
                \begin{tabular}{|l|cccc|c|}
                \hline
                & \multicolumn{4}{|c|}{Target domain} &\\
                \hline
                Methods & \textit{Clp} & \textit{Pnt} & \textit{Rel} & \textit{Skt} & \textit{Avg} \\
                \hline 
                \textit{Source-only}~\cite{DAEL} & $63.44 \pm 0.76$ & $49.92 \pm 0.71$ & $61.54 \pm 0.08$ & $44.12 \pm 0.31$ & $54.76$ \\
                \textit{Oracle}~\cite{DAEL} & $72.59 \pm 0.30$ & $60.53 \pm 0.74$ & $80.47 \pm 0.34$ & $63.44 \pm 0.15$ & $69.26$ \\
                \hline
                DANN~\cite{DANN} & $65.55 \pm 0.34$ & $46.27 \pm 0.71$ & $58.68 \pm 0.64$ & $47.88 \pm 0.54$ & $54.60$ \\
                DCTN~\cite{DCTN} & $62.06 \pm 0.60$ & $48.79 \pm 0.52$ & $58.85 \pm 0.55$ & $48.25 \pm 0.32$ & $54.49$ \\
                MCD~\cite{MCD} & $62.91 \pm 0.67$ & $45.77 \pm 0.45$ & $57.57 \pm 0.33$ & $45.88 \pm 0.67$ & $53.03$ \\
                M$^3$SDA-$\beta$~\cite{M3SDA} & $64.18 \pm 0.27$ & $49.05 \pm 0.16$ & $57.70 \pm 0.24$ & $49.21 \pm 0.34$ & $55.03$ \\
                MME~\cite{MME} & $68.09 \pm 0.16$ & $47.14 \pm 0.32$ & $63.33 \pm 0.16$ & $43.50 \pm 0.47$ & $55.52$ \\
                DAEL~\cite{DAEL} & \textcolor{blue}{$\mathit{69.95 \pm 0.52}$} & \textcolor{red}{$\bm{55.13 \pm 0.78}$} & \textcolor{blue}{$\mathit{66.11 \pm 0.14}$} & \textcolor{blue}{$\mathit{55.72 \pm 0.79}$} & \textcolor{blue}{$\mathit{61.73}$} \\
                Ours & \textcolor{red}{$\bm{71.38 \pm 0.65}$} & \textcolor{blue}{$\mathit{53.76 \pm 0.71}$} & \textcolor{red}{$\bm{66.23 \pm 0.08}$} & \textcolor{red}{$\bm{56.24 \pm 0.67}$} & \textcolor{red}{$\bm{61.90}$} \\
                \hline
                \end{tabular}
                \end{adjustbox}
                \end{center}
                \captionof{table}{Accuracy (\%) on MiniDomainNet (\textit{Clp}: \textit{Clipart}, \textit{Pnt}: \textit{Painting}, \textit{Rel}: \textit{Real}, \textit{Skt}: \textit{Sketch}, \textit{Avg}: \textit{Average}).}
                \label{tab:perfs_mini_domain_net}
            \end{minipage}&
            \begin{minipage}{0.4\textwidth}
                \begin{center}
                \begin{adjustbox}{width=1\textwidth,center}
                \begin{tabular}{|l|cccc|c|}
                \hline
                & \multicolumn{4}{|c|}{Target domain} &\\
                \hline
                Methods & \textit{Art} & \textit{Clp} & \textit{Pct} & \textit{Rel} & \textit{Avg} \\
                \hline 
                \textit{Source-only}~\cite{DARN} & $58.02$ & $57.29$ & $74.26$ & $77.98$ & $66.89$ \\
                \textit{Oracle}~\cite{DARN} & $71.19$ & $79.16$ & $90.66$ & $85.60$ & $81.65$ \\
                \hline
                M$^3$SDA-$\beta$~\cite{M3SDA} & $64.05$ & $62.79$ & $76.21$ & $78.63$ & $70.42$ \\
                SImpAl$_{50}$~\cite{SImpAl} & $70.80$ & $56.30$ & $80.20$ & $81.50$ & $72.20$ \\
                MFSAN~\cite{MFSAN} & \textcolor{red}{$\bm{72.10}$} & $62.00$ & $80.30$ & $81.80$ & $74.10$ \\
                MDAN~\cite{MDAN} & $68.14$ & $67.04$ & $81.03$ & $82.79$ & $74.75$ \\
                MDMN~\cite{MDMN} & $68.67$ & \textcolor{blue}{$\mathit{67.75}$} & $81.37$ & \textcolor{blue}{$\mathit{83.32}$} & $75.28$ \\
                DARN~\cite{DARN} & $\mathit{70.00}$ & \textcolor{red}{$\bm{68.42}$} & \textcolor{blue}{$\mathit{82.75}$} & \textcolor{red}{$\bm{83.88}$} & \textcolor{blue}{$\mathit{76.26}$} \\
                Ours & \textcolor{blue}{$\mathit{71.49}$} & $67.72$ & \textcolor{red}{$\bm{84.19}$} & $82.99$ & \textcolor{red}{$\bm{76.60}$} \\
                \hline
                \end{tabular}
                \end{adjustbox}
                \end{center}
                \captionof{table}{Accuracy (\%) on Office-Home (\textit{Art}: \textit{Art}, \textit{Clp}: \textit{Clipart}, \textit{Pct}: \textit{Product}, \textit{Rel}: \textit{Real-World}, \textit{Avg}: \textit{Average}).}
                \label{tab:perfs_office_home}
            \end{minipage}
        \end{tabular}
    \end{tabular}
\end{center}

\textbf{Source class conditional distributions alignment. } To assess the efficiency of $\mathcal{L}_{ISCL}$ on source class conditional distributions alignment, CMSDA has been trained separately with $\mathcal{L}_{ce}$ and $\mathcal{L}_{ce} + \lambda_1 \mathcal{L}_{ISCL}$ for each (sources, target) possible combination. Then, the Calinski-Harabasz index (CH-index)~\cite{calinski_harabasz_index}, a clustering quality metric, has been computed on the source examples embedding $\bm{h}$.
Using CH-index, we could identify and evaluate if the class conditional distributions from the different source domains are well aligned. For each (sources, target) combination, the CH-indexes with and without $\mathcal{L}_{ISCL}$ are reported on \autoref{fig:abblation_studies}a. As expected, when $\mathcal{L}_{ISCL}$ is used in the final objective, the CH-index increases systematically. This suggests that $\mathcal{L}_{ISCL}$ aligns efficiently source class conditional distributions while keeping discrimative features.

\subsection{Ablation study}
{
    \begin{figure}[htpb]
    \begin{center}
        \begin{tabular}{cccc}
            \begin{tabular}{c}
                \scalebox{0.48}{
                \begin{tabular}{|c|cccc|}
                \hline
                Target domain & \textit{Clp} & \textit{Pnt} & \textit{Rel} & \textit{Skt} \\
                \hline
                 $\mathcal{L}_{ce}$ & $3.76$ & $4.08$ & $3.66$ & $3.97$ \\
                \hline
                 $\mathcal{L}_{ce} + \lambda_1 \mathcal{L}_{ISCL}$& $\bm{4.02}$ & $\bm{4.47}$ & $\bm{3.92}$ & $\bm{4.26}$ \\
                \hline
                \end{tabular}
                }
                \\
                (\textit{a})
                \\
                \scalebox{0.48}{
                \begin{tabular}{|c|cccc|c|}
                \hline
                & \multicolumn{4}{|c|}{Target domain} &\\
                \hline
                Input & \textit{Clp} & \textit{Pnt} & \textit{Rel} & \textit{Skt} & \textit{Avg} \\
                \hline
                $\bm{z}$ & $69.36$ & $53.03$  & $ 66.03 $ & $ 56.04 $ & $ 61.11$ \\
                \hline
                $\bm{h}$ & \bm{$71.38$} & \bm{$53.76$} & \bm{$66.23$} & \bm{$56.24$} & \bm{$61.90$} \\
                \hline
                \end{tabular}%
                }
                \\
                (\textit{b})
                \\
                \scalebox{0.48}{
                \begin{tabular}{|c|cccc|c|}
                \hline
                & \multicolumn{4}{|c|}{Target domain} &\\
                \hline
                 Mixup & \textit{Clp} & \textit{Pnt} & \textit{Rel} & \textit{Skt} & \textit{Avg} \\
                \hline 
                w/o Mixup & $70.32$ & $51.75$ & $64.92$ & $51.11$ & $59.52$ \\
                \hline
                w Mixup & $\bm{71.38}$ & $\bm{53.76}$  & $\bm{66.23}$ & $\bm{ 56.24 }$ & $\bm{ 61.90 }$ \\
                \hline
                \end{tabular}%
                }
                \\
                (\textit{c})
            \end{tabular}
            & \multicolumn{3}{c}{
                \begin{tabular}{c}
                    \includegraphics[width=0.70\textwidth]{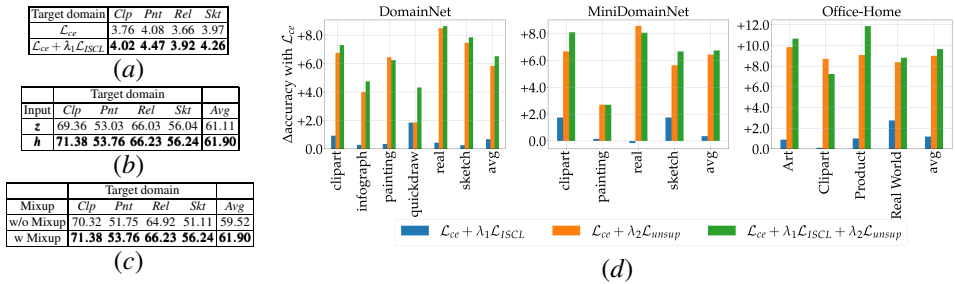} \\
                    (\textit{d})
                \end{tabular}
            }
        \end{tabular}
    \end{center}
    \caption{CH indexes for each combination of target and sources domains of MiniDomainNet considering only the source embeddings $\bm{h}$ with or without $\mathcal{L}_{ISCL}$ (\textit{a}), performances in terms of accuracy on MiniDomainNet with classification on $\bm{h}$ or $\bm{z}$ embeddings (\textit{b}), with or without MixUp and interpolated losses (\textit{c}), $\Delta$accuracy compared to standard $\mathcal{L}_{ce}$ for different combination of losses (\textit{d}).}
    \label{fig:abblation_studies}
    \end{figure}
}

\textbf{Classification on $\bm{h}$ or $\bm{z}$. }Similarly to self-supervised methods such as \cite{SimCLR}, we have investigated the behavior of our model when the classification head takes as input the features extractor representation $\bm{h}$ or the projection head representation $\bm{z}$. For each target domain of MiniDomainNet, the framework has been trained by either feeding $\bm{h}$ or $\bm{z}$ to the classification head. The accuracies obtained on the different target domains of MiniDomainNet are reported in \autoref{fig:abblation_studies}b. Using the representation $\bm{z}$ instead of $\bm{h}$ leads to little lower performances on every target domain but performances remain competitive to other multi-source UDA methods. For the \textit{Clipart} domain, using $\bm{z}$ results in a $2\%$ accuracy drop. 
These performances discrepancies are in adequacy with the findings of \cite{SimCLR} arguing that $\bm{h}$ usually provides better representations for the final classification task. Therefore, even if performances are quite comparable, we recommend feeding  $\bm{h}$ to the classification head $C$.

\textbf{MixUp and interpolated losses. } To assess the contributions of MixUp and interpolated losses on CMSDA performances,  we have trained two different versions of the framework. In the first version, MixUp is removed while standard versions of the cross entropy and supervised contrastive losses are used. Conversely, in the second version, MixUp is applied on source examples and interpolated versions of the cross entropy and the supervised contrastive losses are used. Accuracies for these two versions and for each target domain of MiniDomainNet are reported on \autoref{fig:abblation_studies}c. It suggests that MixUp combined with the interpolated losses brings a significant gain of performances. The average accuracy gain is more than $2\%$. We believe that MixUp applied on examples from different source domains can be seen as a way to mix source domains style which serves as an efficient data augmentation to learn domain invariant features. Additionally, we think that the combination of MixUp and interpolated cross entropy improves the model calibration and robustness on out-of-distribution data. This enables cleaner pseudo-labels for target examples.

\textbf{Loss ablation.} 
To highlight the influence of each loss on the overall performances, CMSDA has been trained with different combinations of the losses $\mathcal{L}_{ce}$, $\mathcal{L}_{ISCL}$ and $\mathcal{L}_{unsup}$. For this experiment, the hyperparameters remain unchanged. For each dataset and each combination of losses, we report on \autoref{fig:abblation_studies}d the gain/loss in terms of accuracy compared to minimizing only $\mathcal{L}_{ce}$. When $\mathcal{L}_{ISCL}$ is combined with $\mathcal{L}_{ce}$ (blue bars), it has in average a positive impact  ($0.68\%$, $0.87\%$ and $1.19\%$ respectively for DomainNet (DN), MiniDomainNet (MDN) and OfficeHome (OH)). More specifically, it brings significant gains for target domains with large domain shift (DN-\textit{quickdraw}: $+1.85\%$, MDN-\textit{sketch}:$+1.75\%$ or OH-\textit{Real World}: $+2.75\%$). Even if in some rare cases, such as MDN-\textit{real}, $\mathcal{L}_{ISCL}$ leads to a very small loss of accuracy, its overall contribution is beneficial. When $\mathcal{L}_{unsup}$ is added to $\mathcal{L}_{ce}$ (orange bars), performances are systematically improved. In general, $\mathcal{L}_{unsup}$ contribution is higher than $\mathcal{L}_{ISCL}$. This is consistent with the performances gap usually observed between methods that do exploit target data (multi-source UDA) and the ones that do not (multi-source Domain Generalization).
Additionally, when $\mathcal{L}_{ISCL}$ is combined to $\mathcal{L}_{ce}$ and $\mathcal{L}_{unsup}$ (green bars), the performances are often enhanced and especially on the most challenging target domains (DN-\textit{quickdraw}: $+2.48\%$, DN-\textit{infograph}: $+0.75\%$, MDN-\textit{sketch}: $+1.01\%$). The accuracy gains of $\mathcal{L}_{ISCL}$ and $\mathcal{L}_{unsup}$ seem to be additive suggesting that their contributions are independent. These observations prove the usefulness of each loss and their independent contributions to the overall framework.
\subsection{Sensitivity to hyperparameters}
    \begin{figure}[htpb]
    \begin{center}
    \begin{tabular}{cccc}
         \includegraphics[width=0.24\textwidth]{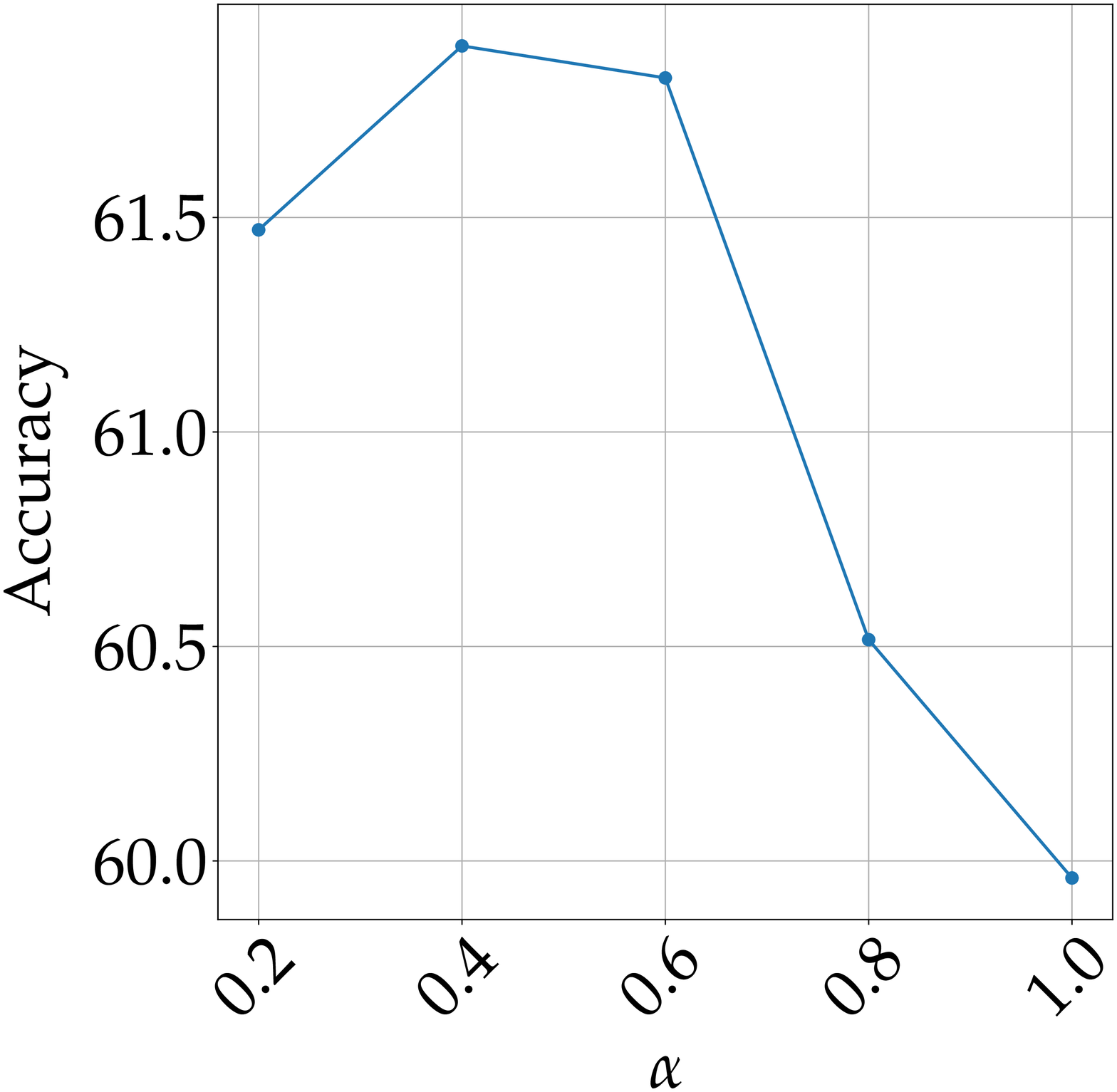}&
         \includegraphics[width=0.24\textwidth]{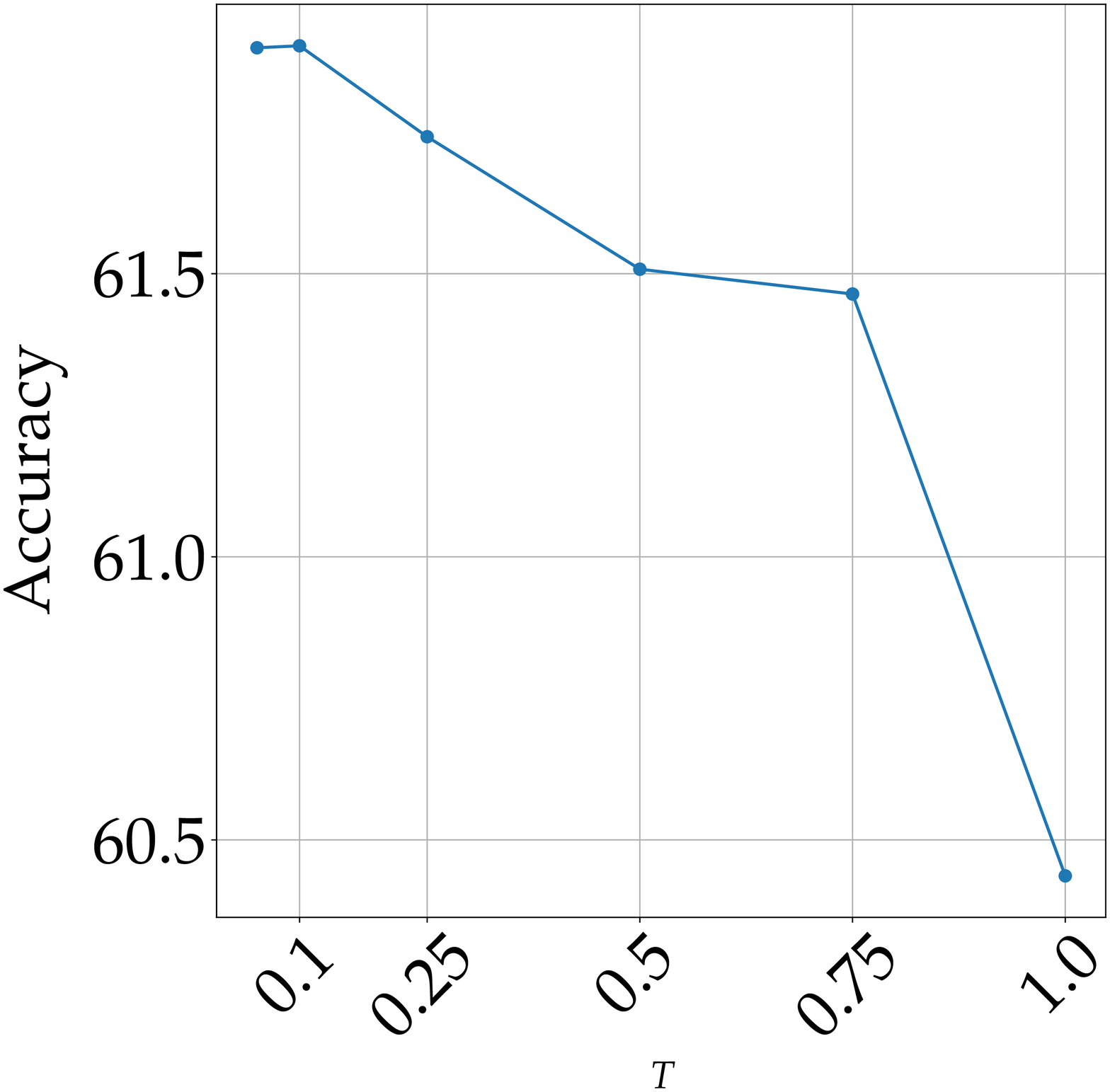}&
         \includegraphics[width=0.24\textwidth]{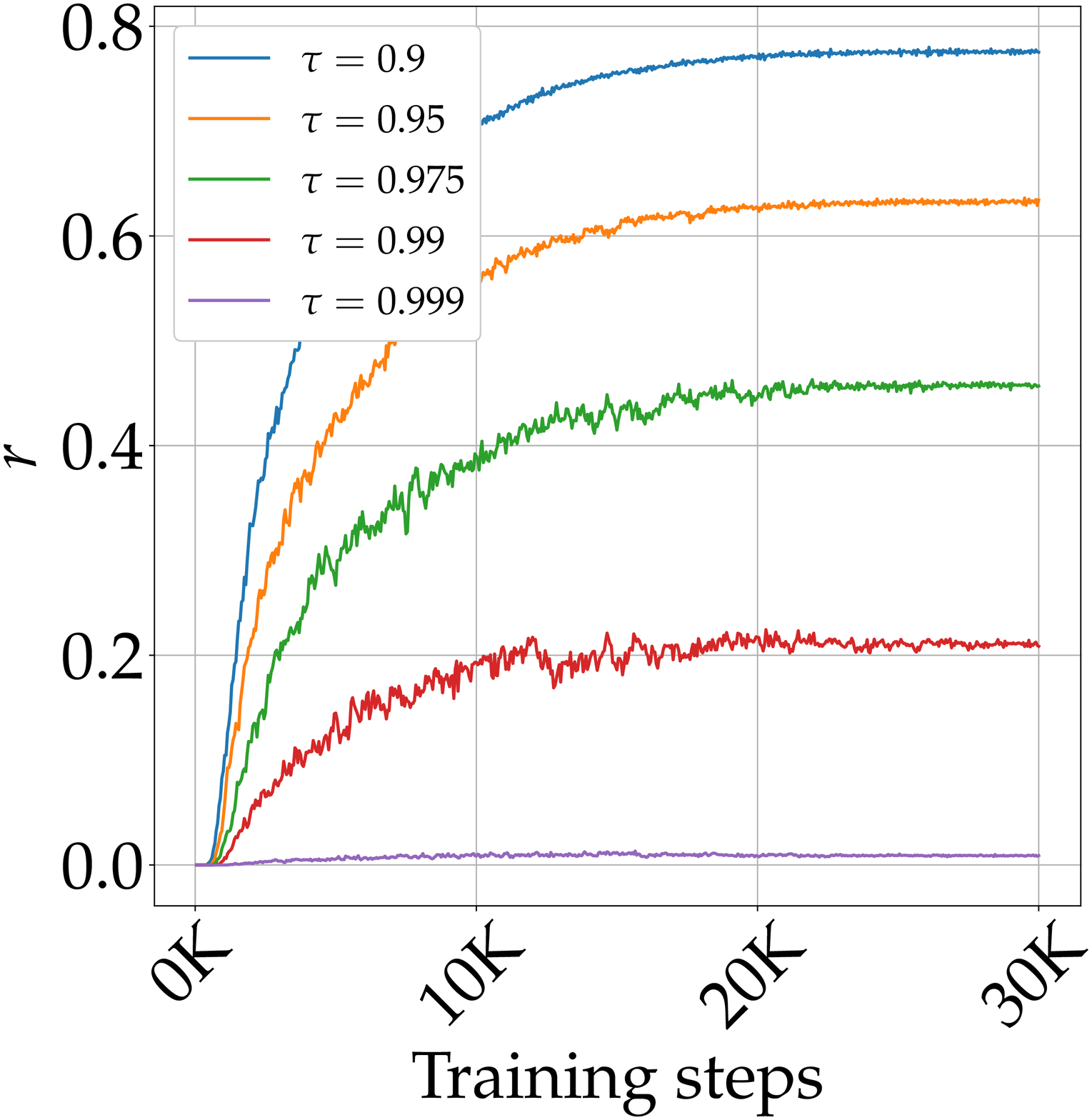}&
         \includegraphics[width=0.24\textwidth]{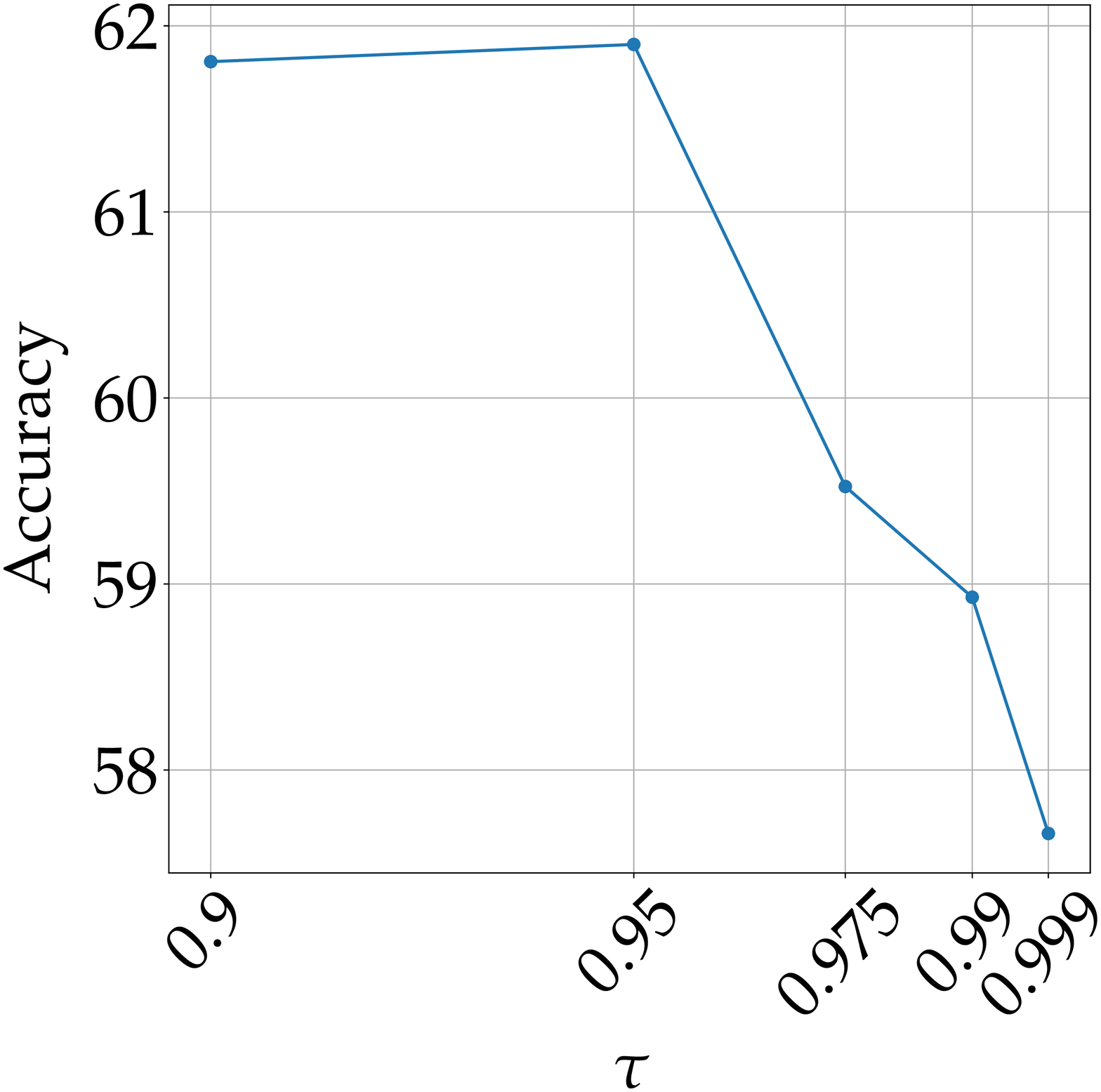}\\
         (\textit{a}) & (\textit{b}) & (\textit{c}) & (\textit{d})\\
    \end{tabular}
    \end{center}
    \caption{Averaged accuracy (\%) on MiniDomainNet with respect to $\alpha$~ (\textit{a}); $T$ (\textit{b}); $\tau$ (\textit{d}). Ratio $r$ of target examples contributing to $\mathcal{L}_{unsup}$ with respect to $\tau$ (\textit{c}).}
    \label{fig:hyperparameters_sensitivity}
    \end{figure}
In this section, we assess the behavior of the model with respect to its hyperparameters $\alpha$, $T$, and $\tau$. Experiments about $\lambda_1$ and $\lambda_2$ are included in the supplementary material. 

\textbf{MixUp hyperparameter $\alpha$. } 
In Mixup \cite{mixup_training}, the interpolation parameter $\lambda \in [0, 1]$ is drawn such that $\lambda \sim Beta (\alpha, \alpha)$. $\alpha$ controls the interpolation strength. 
Too low $\alpha$ usually lead to weak regularization while too high $\alpha$ lead to too strong regularization resulting in underfitting and under-confident models \cite{mixup_training, mixup_calibration}. 
We have investigated how $\alpha$ impacts the performances by training the framework with different $\alpha$ and for each target domain of MiniDomainNet. The average accuracy over the target domains with respect to $\alpha$ is reported on \autoref{fig:hyperparameters_sensitivity}a.
Performances are quite stable for $\alpha \in [0.2, 0.6]$ but start decreasing for $\alpha > 0.6$, validating the observations made in ~\cite{mixup_training, mixup_calibration}. $\alpha=0.4$ leads to the best accuracy. To work on other datasets, we suggest $\alpha=0.4$ as a starting point.

\textbf{Temperature $T$. }
The temperature hyperparameter $T$ is known to have a crucial role in self-supervised/supervised contrastive learning methods~\cite{SimCLR, supervised_contrastive_learning}. Setting $T$ properly can result in a non negligible gain of performances \cite{supervised_contrastive_learning}. 
According to \cite{contrastive_loss_behavior}, selecting the optimal $T$ is a compromise between uniformity and tolerance. Uniformity corresponds to the capacity of the representations $\bm{\tilde{z}}$ to be uniformly distributed over the sphere while tolerance describes how close the representations are for examples in the same class. Uniformity has known to be important to learn separable features however high uniformity induces a decrease of tolerance \cite{contrastive_loss_behavior}. $T \rightarrow 0$ tends to encourage uniformity whereas $T \rightarrow +\infty$ promotes tolerance. 
To evaluate the effect of the temperature $T$ on the performances, we have trained our model with different $T$ and for each target domain of MiniDomainNet. The average accuracy over the target domains with respect to $T$ is reported on \autoref{fig:hyperparameters_sensitivity}b. It seems that our framework benefits from lower temperatures. Indeed, for $T \in [0.05, 0.1]$, the performances are quite stable and reach the maximum for $T=0.1$. For $T > 0.1$, the accuracy begins to decrease slowly until $T=0.75$ and then it drops. Overall, for multi source UDA, our experiments suggest that uniformity is more important than tolerance however too low temperatures ($T < 0.05$) might as well impact negatively the model performances. We suggest to use a temperature $T=0.1$ for further experiments on different datasets.

\textbf{FixMatch probability threshold $\tau$. }
In semi-supervised problems, including false pseudo-labeled examples into the training could lead to lower performance~\cite{pseudo_labeling}. In $\mathcal{L}_{unsup}$, $\tau$ addresses this problem by discarding non confident pseudo-labeled examples. More specifically, pseudo-labeled examples whose maximum predicted probability is below $\tau$ do not contribute to $\mathcal{L}_{unsup}$.
To illustrate the trade off between the number of target pseudo labels used by CMSDA, on \autoref{fig:hyperparameters_sensitivity}c and for the target domain \textit{Clipart} of MiniDomainNet, we have plotted for different values of $\tau$ the ratio of target examples $r$ that contribute to the loss $\mathcal{L}_{unsup}$. $r$ is defined as follows:
\begin{equation}
    r = \dfrac{\displaystyle \sum_{i=1}^{N_T} \mathbbm{1}_{\{\max \bm{\hat{y}_i^{(w)}} > \tau\}}}{N_T}
\end{equation}
\autoref{fig:hyperparameters_sensitivity}c reveals that for any value $\tau$, as the training progresses, the model gets more and more confident predictions resulting in an increase of $r$. A second observation is that when $\tau$ is set too low ($\tau = 0.9$), $r$ is in average high at the end of training $(r \sim 77\%)$ whereas \textit{Oracle} reaches only $72.59\%$ accuracy. This suggests that false pseudo-labeled target examples contributes to $\mathcal{L}_{unsup}$. On the contrary, when $\tau$ is set too high ($\tau = 0.999$), $r$ is in average very low at the end of training $(r \simeq 0.02)$. This indicates that a lot of correct pseudo-labeled target examples have been discarded. To evaluate the effect of $\tau$ on the performances, CMSDA has been trained with different $\tau$ values for each target domain of MiniDomainNet, while the rest of hyperparameters remain unchanged. The average accuracy over target domains with respect to $\tau$ is reported in \autoref{fig:hyperparameters_sensitivity}d. Performances are stable when choosing $\tau \in [0.9, 0.95]$. $\tau = 0.95$ leads to the highest average accuracy. For $\tau < 0.95$, false pseudo labels contribute to $\mathcal{L}_{unsup}$ resulting in a small drop of accuracy. For $\tau > 0.95$, as $\tau$ increases, more and more correct pseudo-labeled examples are discarded and performances start to drop.

\section{Conclusion}
In this work, we have introduced a new framework combining recent advances in supervised contrastive learning and semi-supervised learning to address the problem of multi-source UDA. Our framework, through supervised contrastive learning, is able to align source class conditional distributions resulting in more robust and universal features for the target domain. Simultaneously, the model adjusts to the target domain via hard pseudo labeling and consistency regularization on target examples.
Our framework has been evaluated on $3$ datasets commonly used for multi-source UDA and has reported superior results over previous state-of-the-art methods with robust results on complex domain where negative transfer can occur. 

In future research, we plan to explore the use of supervised contrastive learning on both source and pseudo-labeled target examples  so as to align source and target conditional distributions all together. Additionally, we believe that data augmentation strategies usually designed for domain generalization (MixStyle~\cite{mixstyle}, Fourier Based Augmentation~\cite{fourier_based_augmentation}) and conditional normalizations could provide interesting directions for our future work.


\bibliography{biblio.bib}
\end{document}


\maketitle

\section{Datasets details}

\textbf{DomainNet. } 
DomainNet is the largest dataset for benchmarking domain adaptation methods. It has been introduced by~\cite{M3SDA} and contains more than $0.6$ millions images for six distincts domains (\textit{clipart, infograph, painting, quickdraw, real, sketch}) spread over 345 classes.

\textbf{MiniDomainNet. }
MiniDomainNet~\cite{DAEL} is a subset of DomainNet that uses less images ($\sim$ $140K$ images) with smaller size of $96 \times 96$ spread over $4$ selected domains and $126$ selected classes. MiniDomainNet was introduced to reduce the requirements for computing resources and to remove noisy domains/examples. Therefore, MiniDomainNet will also be used for our ablation studies.

\textbf{Office-Home. }
Office-home~\cite{office_home} has been widely used as a standard dataset for evaluating domain adaptation methods. It includes approximately $15500$ images from 4 different domains: \textit{Art}, \textit{Clipart}, \textit{Product} and \textit{Real-World}. For each domain, it contains images of 65 object categories found typically in Office and Home settings.

\section{Implementation details}

\textbf{Strong data augmentation. }Strong data augmentation $t_s(.)$ is based on the same image transformations as RandAugment \cite{randaugment}.
An image is augmented with $t_s(.)$ following these steps:
\begin{enumerate}
    \item Two transformations are sampled randomly in the list of possible transformations (\autoref{tab:rand_augment}) and the parameters defining these transformations are drawn randomly inside their corresponding range (\autoref{tab:rand_augment}).
    \item The image is horizontally and randomly flipped with a probability $p=0.5$.
    \item Random cropping is applied to the image with a crop size between $[90\%, 100\%]$ of the original image size. The crop is resized to $(96, 96)$ for MiniDomainNet, $(180, 180)$ for DomainNet and $(224, 224)$ for Office-Home. 
\end{enumerate}

\textbf{Weak data augmentation. }Weak data augmentation $t_w(.)$ is made only of random horizontal flips. The image is finally resized to $(96, 96)$ for MiniDomainNet, $(180, 180)$ for DomainNet and $(224, 224)$ for Office-Home.

\begin{table}[t!]
    \begin{center}
        \begin{adjustbox}{width=0.9\textwidth,center}
        \begin{tabular}{|L{2.2cm}|C{2.2cm}|C{1.5cm}|L{8cm}|}
            \hline
            Transformation & Hyperparameter & Range & Description \\
            \hline
            AutoContrast & $C$ & $[0, 1]$ & Maximize (normalize) image contrast. It calculates a histogram of the input image, removes $C$ percent of the lightest and darkest pixels from the histogram, and remaps the image so that the darkest pixel becomes black (0), and the lightest becomes white (255).\\
            \hline
            Brightness & $B$ & $[0.1, 1.9]$ & Control image brightness by a factor $B$. $B = 0$ gives a black image and $B=1$ gives the original image.\\
            \hline
            Color & $C$ & $[0.1, 1.9]$ & Adjust the colour balance of an image given a enhancement factor $C$. $C = 0$ gives a black and white image, $C=1$ gives the original image. \\
            \hline
            Contrast & $C$ & $[0.1, 1.9]$ & Control the contrast of an image given an enhancement factor $C$. $C=0$ gives a solid grey image and $C=1$ gives the original image.\\
            \hline
            Equalize & & $[0, 1]$ & Equalize the image histogram. \\
            \hline
            Identity & & $[0, 1]$ & Returns the original image.\\
            \hline
            Invert & & $[0, 1]$ & Invert the image.\\
            \hline
            Posterize &$B$& $[4, 8]$ & Reduce the number of bits to $B$ bits for each color channel.\\
            \hline
            Rotate & $\theta$ & $[-30, 30]$ & Rotate the image counter clocksize with an angle $\theta$. \\
            \hline
            Sharpness & $S$ & $[0.1, 1.9]$ & Adjust the image sharpness given an enhancement factor $S$. $S=0$ gives a blurred image, $S=1$ the original image and $S=2$ an sharpened image.\\
            \hline
            ShearX & $R$ & $[-0.3, 0.3]$ & Shear the image along the horizontal axis with rate $R$\\
            \hline
            ShearY & $R$ & $[-0.3, 0.3]$ & Shear the image along the vertical axis with rate $R$\\
            \hline
            Solarize & $S$ & $[0, 256]$ & Invert all pixel values above a threshold $S$.\\
            \hline
            TranslateX & $t$ & $[-0.3, 0.3]$ & Translate an image with size $(H, W)$ along the horizontal axis by $t \times W$ pixels.\\
            \hline
            TranslateY & $t$ & $[-0.3, 0.3]$ & Translate an image with size $(H, W)$ along the vertical axis by $t \times H$ pixels.\\
            \hline
        \end{tabular}
        \end{adjustbox}
    \end{center}
    \caption{Transformations used for strong data augmentation $t_s(.)$. Most of the transformations are described by some parameters. When one transformation depending on parameters is drawn, parameters are randomly sampled inside the specified Range.}
    \label{tab:rand_augment}
\end{table}

\textbf{Architecture and hyperparameters. }
For DomainNet and Office-Home, the features extractor $F$ corresponds to a ResNet50~\cite{resnet} while for MiniDomainNet a ResNet18 is used. All networks are pretrained on the ImageNet dataset~\cite{imagenet}.
The embeddings $\bm{h}$ of the ResNet50 and the ResNet18 are respectively the $2048$ dimensional features vector ($d_1 = 2048$) and the $512$ dimensional features vector ($d_1 = 512$) obtained after the global average pooling layer. In all experiments, dimension of the projection head representation $\bm{z}$ is set to $d_2 = 256$, temperature $T$ is set to $0.1$ and probability threshold $\tau$ to $0.95$. The $\alpha$ hyperparameter for MixUp is set to $0.4$ for DomainNet, MiniDomainNet and $0.2$ for Office-Home. $\lambda_{1}$ and $\lambda_{2}$ are respectively set to $0.1$ and $1$. Details about the strong $t_s(.)$ and weak $t_w(.)$ data augmentations are presented in the supplementary material.
All our experiments are conducted using the PyTorch library~\cite{pytorch} with $2$ Nvidia Tesla V100 GPUs. For optimization, Adam~\cite{adam} method is used with an initial learning rate of $10^{-4}$ and a cosine decay learning rate with a minimum learning rate of $0$. On DomainNet, MiniDomainNet and Office-Home, the models are trained respectively for $70K$, $60K$ and $20K$ iterations. For these datasets, source and target batch sizes $(N_S, N_T)$ are respectively set to $(256, 256)$, $(256, 256)$ and $(128, 128)$. For each raw source example, two strongly augmented examples are generated.

\textbf{Target examples pseudo labeling. }

An exponential moving average model, as in the original FixMatch method \cite{FixMatch}, is used to obtain the pseudo labels on the weakly augmented target examples. This model is also used for the final evaluation on the target domain. For all the experiments in the main paper or the supplementary material, exponential decay parameter is set to $\alpha_{EMA} = 0.999$.

\section{Extensive experiments}

\begin{figure}[h]
\begin{center}
\begin{tabular}{cc}
     \includegraphics[width=0.3\textwidth]{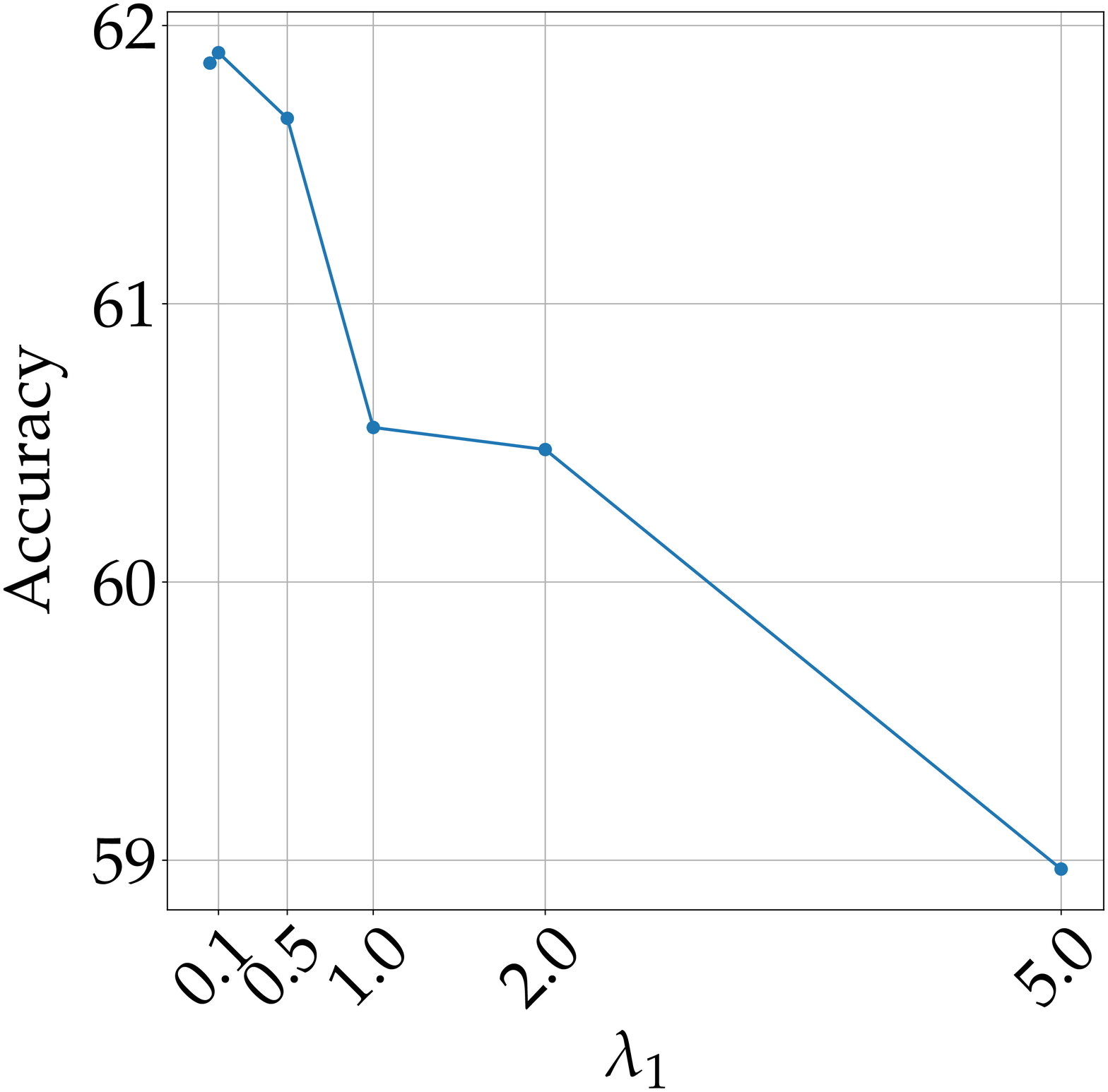}&
     \includegraphics[width=0.3\textwidth]{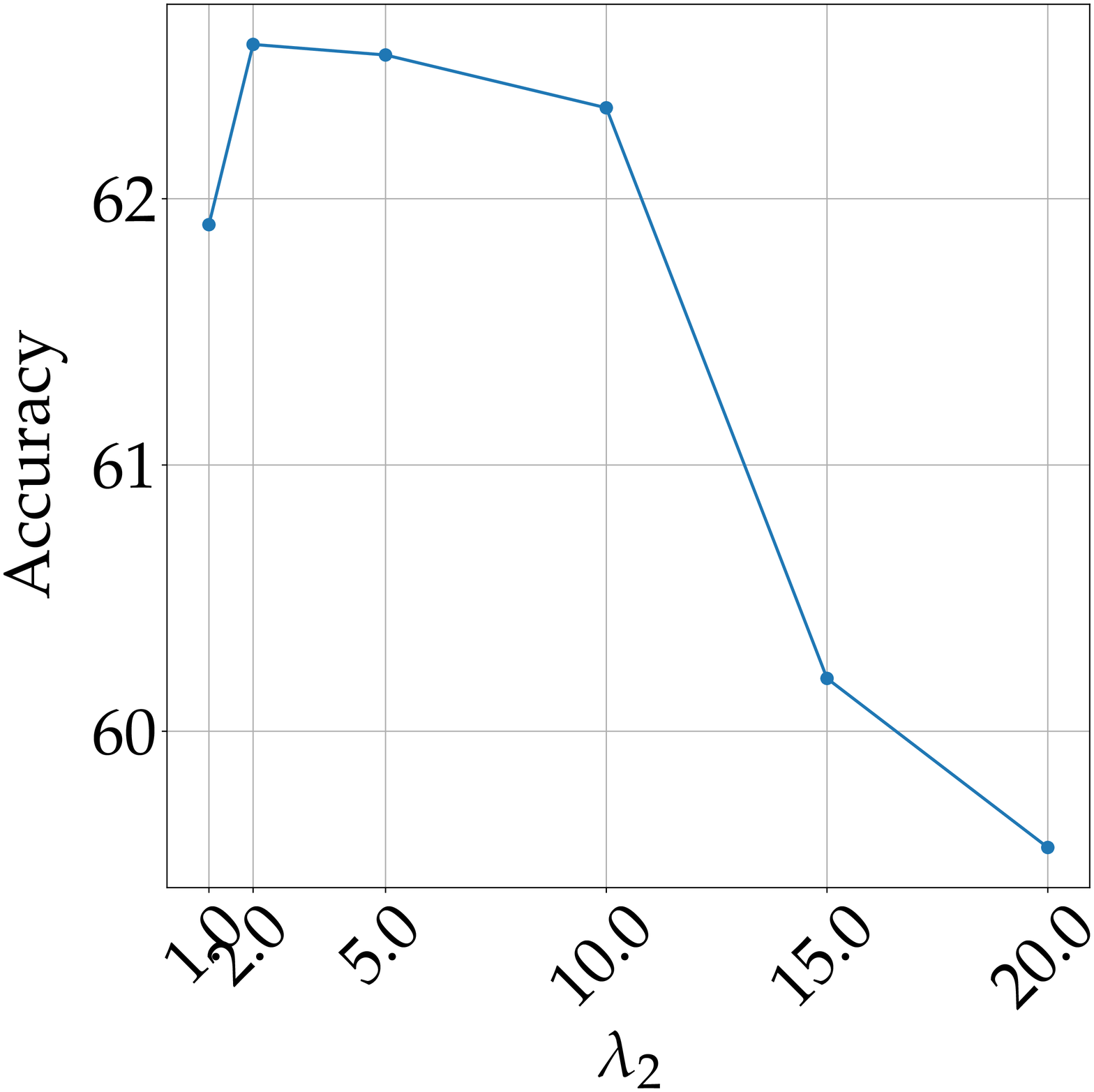}\\
     (\textit{a}) & (\textit{b})
\end{tabular}
\end{center}
\caption{Averaged accuracy on MiniDomainNet with respect to $\lambda_1$(\textit{a}) and $\lambda_2$(\textit{b}).}
\label{fig:hyperparameters_sensitivity}
\end{figure}
\textbf{Performances with respect to $\lambda_{1}$ and $\lambda_{2}$. }$\lambda_{1}$ and $\lambda_{2}$ are losses hyperparameters weighting respectively $\mathcal{L}_{ISCL}$ and $\mathcal{L}_{unsup}$ in the final objective. Setting $\lambda_1$ and $\lambda_2$ is a compromise between
giving enough importance to $\mathcal{L}_{ISCL}$ to align source class conditional distributions enabling general and transferable features learning and giving enough importance to $\mathcal{L}_{unsup}$ to adjust features for the target domain.

To find a reasonable starting combination of $(\lambda_1, \lambda_2)$ for further experiments and assess their respective influence on the model performances, CMSDA has been trained for each target domain of MiniDomainNet with two different settings. In the first setting, CMSDA is trained with different $\lambda_1$ while setting $\lambda_2=1$. In the second setting, CMSDA is trained with different $\lambda_2$ while fixing $\lambda_1=0.1$. Average accuracies over MiniDomainNet target domains are reported for each experiment on \autoref{fig:hyperparameters_sensitivity}a and \autoref{fig:hyperparameters_sensitivity}b. From \autoref{fig:hyperparameters_sensitivity}a, we can notice that performances remain stable when $\lambda_1 \in [0.05, 0.5]$ but still the best accuracy is achieved for $\lambda_1=0.1$. For $\lambda_1 > 0.5$, performances starts to decrease quickly.
From \autoref{fig:hyperparameters_sensitivity}b, it seems that performances remains stable for a large range of $\lambda_2$ ($\lambda_2 \in [2, 10]$). The best average accuracy is achieved for $\lambda_2=2$ while for $\lambda_2 < 2$ and $\lambda_2 > 10$ performances begin to decline. For experimenting CMSDA on other datasets, we suggest starting with $(\lambda_1, \lambda_2) = (0.1, 2)$.

\textbf{Source examples mixing strategies. }
A comparison between different mixing augmentation strategies for CMSDA framework is presented in this section. More specifically, we have compared the performances of our framework when MixUp \cite{mixup_training} or CutMix \cite{cutmix} is implemented. The average on each target domain of MiniDomainNet and for the two different mixing strategies are reported on \autoref{tab:mixing_strategies}.
\begin{table}[htpb]
    \begin{center}
        \begin{tabular}{|l|c|c|}
        \hline
        {} &   MixUp &  CutMix \\
        \hline
        \textit{clipart}  &  71.38 &   69.68 \\
        \hline
        \textit{painting} &  53.76 &   53.18 \\
        \hline
        \textit{real}     &  66.23 &   67.14 \\
        \hline
        \textit{sketch}   &  56.24 &   51.43 \\
        \hline
        \textit{average}      &  61.90 &   60.36 \\
        \hline
        \end{tabular}
    \end{center}
    \caption{Performances on MiniDomainNet when using different mixing strategies.}
    \label{tab:mixing_strategies}
\end{table}
These results reveal that CutMix has comparable performances with MixUp except on the domain \textit{sketch} where MixUp outperforms CutMix. We believe that by fine-tuning the CutMix hyperparameters, it would be possible to close the small performances gap between MixUp and CutMix. However, overall, there is no clear advantages to use one mixing method over another.

\bibliography{biblio}